# Towards Accessible Learning: Deep Learning-Based Potential Dysgraphia Detection and OCR for Potentially Dysgraphic Handwriting

Dr Vydeki D[1], Divyansh Bhandari[1], Pranav Pratap Patil[1], Aarush Anand Kulkarni[1]

[1]School of Electronics Engineering

Vellore Institute of Technology, Chennai

## Abstract –

Dysgraphia is a learning disorder that affects handwriting abilities, making it challenging for children to write legibly and consistently. Early detection and monitoring are crucial for providing timely support and interventions. This study applies deep learning techniques to address the dual tasks of dysgraphia detection and optical character recognition (OCR) on handwriting samples from children with potential dysgraphic symptoms. Using a dataset of handwritten samples from Malaysian schoolchildren, we developed a custom Convolutional Neural Network (CNN) model, alongside VGG16 and ResNet50, to classify handwriting as dysgraphic or non-dysgraphic. The custom CNN model outperformed the pre-trained models, achieving a test accuracy of 91.8% with high precision, recall, and AUC, demonstrating its robustness in identifying dysgraphic handwriting features. Additionally, an OCR pipeline was created to segment and recognize individual characters in dysgraphic handwriting, achieving a character recognition accuracy of approximately 43.5%. This research highlights the potential of deep learning in supporting dysgraphia assessment, laying a foundation for tools that could assist educators and clinicians in identifying dysgraphia and tracking handwriting progress over time. The findings contribute to advancements in assistive technologies for learning disabilities, offering hope for more accessible and accurate diagnostic tools in educational and clinical settings.

## 1. INTRODUCTION

Dysgraphia, a specific learning disability affecting writing skills, is characterized by difficulties in handwriting, spelling, and organizing written text. It impacts a significant number of children globally, affecting their academic performance and self-esteem. The ability to effectively communicate through writing is a critical component of educational development, and children with dysgraphia often face significant barriers in this area. Early and accurate detection of dysgraphia is crucial for timely intervention, which can help mitigate its negative effects on children's educational progress and psychological well-being. However, traditional methods of diagnosing dysgraphia often rely on subjective clinical assessments, which can be inconsistent, time-consuming, and inaccessible for many families. Furthermore, interpreting the handwriting of dysgraphic children presents additional challenges, as it is often irregular and difficult to understand. These challenges underscore the need for more efficient, objective, and scalable diagnostic and interpretive solutions.

Recent advancements in technology, particularly in the field of deep learning, have shown significant promise in automating the detection of learning disabilities like dysgraphia and addressing the difficulties in interpreting dysgraphic handwriting. By leveraging machine learning models to analyze handwriting features such as stroke patterns, pressure, and writing speed, it becomes possible to identify unique characteristics associated with dysgraphia. In addition, specialized OCR (Optical Character Recognition) systems can be developed to assist in interpreting dysgraphic handwriting, thereby providing support not only in diagnosing dysgraphia but also in bridging communication gaps between affected children and their educators or caregivers.

The primary objective of this research is to address both the detection and interpretation of dysgraphia. The first goal is to develop a deep learning-based model capable of classifying children as either dysgraphic or non-dysgraphic based on their handwriting samples. The study utilizes various deep learning algorithms, including Convolutional Neural Networks (CNNs), ResNet, VGG, and other architectures, to extract both spatial and temporal features from digital handwriting data. By conducting a comparative study of these different models, we aim to determine which architecture is most effective at recognizing dysgraphic handwriting and provides the highest accuracy. This comparative analysis will help identify the strengths and limitations of each model, ultimately leading to an objective and data-driven approach for dysgraphia detection.

Another key aspect of this research is the development of a specialized OCR system that can assist in interpreting dysgraphic handwriting. Unlike traditional OCR systems, which are designed for well-formed text, this OCR model is tailored to recognize the irregular and often inconsistent handwriting patterns of dysgraphic children. The system incorporates a specialized library that analyzes handwritten words, checks them against a language-specific dictionary, and provides corrected versions of the words or sentences. This feature is intended to support parents and teachers in understanding what the child has written, bridging the gap between the child's intended message and its readability. By offering an accurate interpretation of dysgraphic handwriting, this tool not only aids in diagnosis but also enhances communication, thereby providing practical support to dysgraphic children, their families, and educators.

The proposed system aims to make significant progress in the field of dysgraphia research by integrating deep learning-based classification with an innovative OCR solution tailored to the unique challenges of dysgraphic handwriting. By developing accessible and scalable diagnostic tools, this research contributes to the broader goal of ensuring that children with dysgraphia receive the timely support they need to succeed academically and socially.

## 2. LITERATURE REVIEW

Machine learning has become an effective tool for identifying dysgraphia in a scalable and objective way, moving beyond the limitations of traditional clinical assessments. Recent research, including Drotár and Dobeš (2020) [1], has leveraged digital tablets to capture nuanced handwriting data, such as writing speed, pressure, and pen lifts, using algorithms like AdaBoost, random forests, and SVM to reach nearly 80% accuracy. Studies led by Asselborn et al. (2019) [25] have shown that classifiers like random forests and neural networks, along with clustering techniques, can categorize dysgraphia by analyzing specific handwriting features. Nonetheless, challenges like small sample sizes, age-related variations, and orthographic differences remain, affecting the models' generalizability. The field recognizes the need for larger, more diverse datasets to improve model precision and reliability. Despite these challenges, machine learning continues to advance automated handwriting analysis for dysgraphia diagnosis, potentially leading to accessible, early interventions for children affected by this condition.

In recent studies, machine learning models have been employed to analyze graphomotor tasks recorded with digital tablets to assist in the early detection of dysgraphia in children (Devillaine et al., 2021) [2]. Traditional assessments, such as the BHK test, often involve subjective judgment, language dependence, and delayed testing until children have practiced writing for two years. Consequently, researchers have investigated drawing-based graphomotor tasks, capturing kinematic data like stroke duration, pen lifts, and path deviation to differentiate dysgraphic children from their peers. For instance, Devillaine et al. [2] created a model from the graphomotor data of 305 children, achieving a classification accuracy of 73% by focusing on variations in children's drawing patterns. Although there are challenges, such as pen pressure calibration issues and differentiating between handwriting and drawing tasks, this approach suggests the potential for language-independent dysgraphia pre-diagnosis, which could offer a scalable and objective diagnostic tool for early intervention.

Machine learning and deep learning approaches are increasingly utilized to detect learning disabilities like dyslexia and dysgraphia. Traditional dysgraphia assessments rely on subjective human analysis, making them inconsistent and dependent on specialist availability. Studies by Asselborn et al. (2019) [25] have shown that feature extraction from handwriting, combined with classifiers such as random forests, can accurately identify dysgraphia. However, most research has focused on European languages, leaving a gap for languages with unique orthographic characteristics, such as Hindi. Addressing this, Yogarajah and Bhushan (2020) [3] developed a CNN model specifically for Hindi handwriting, incorporating features unique to the language, like matras and conjoined consonants. Their model reached an accuracy of around 86%, showcasing the effectiveness of deep learning in dyslexia-dysgraphia detection for Hindi. This underscores the need for language-specific machine learning adaptations to improve diagnostic accessibility and accuracy across diverse linguistic contexts.

Studies on dysgraphia detection increasingly focus on using machine learning to enhance diagnostic accuracy and accessibility. Traditional methods like the BHK test depend on subjective judgment and often overlook dynamic handwriting features, leading to potential inconsistencies. Recent research has applied machine learning algorithms, including SVMs, to analyze kinematic handwriting data, aiming for a more objective dysgraphia diagnosis. Prior work by Mekyska et al. and Asselborn et al. (2019) [25] utilized metrics such as pen pressure and tilt but faced limitations due to small sample sizes and device constraints. A study by Deschamps et al. (2021) [4] addressed these limitations by using a large dataset of 580 children, collected across various tablets and software, with the goal of creating a scalable, device-independent diagnostic tool. The inclusion of children with varying dysgraphia severity levels enhances the model's robustness. By employing moving z-score standardization to account for age differences, this study seeks to improve diagnostic accuracy, advancing automated tools that support early intervention.

Machine learning techniques are being explored to address the challenges of diagnosing learning disabilities like dyslexia and dysgraphia, which affect a significant portion of the global population and have long-term impacts on academic and social outcomes. Traditional diagnosis requires extensive, subjective assessments by specialists, which can be costly and time-consuming. Recent research aims to automate this process by training machine learning models on specific handwriting and reading features. For instance, Richard and

Serrurier (2020) [5] used audio recordings of word reading and images of handwriting to analyze reading errors, word retracing, and handwriting features such as letter size and spacing. Preliminary models in these studies have shown that machine learning can reliably distinguish between typical and dyslexic or dysgraphic individuals. Techniques like random forests and SVM classifiers are commonly applied, and recent advances suggest these models could serve as efficient, non-invasive screening tools. However, data diversity and feature stability are essential, as researchers continue to seek broader datasets to enhance model accuracy and ensure applicability.

Literature on automated dysgraphia diagnosis highlights the potential of machine learning to create scalable, efficient tools for early intervention. Traditional assessments are often subjective and inaccessible, emphasizing the need for objective and standardized approaches. Recent studies focus on automated systems that capture dynamic handwriting features, such as pen pressure, tilt, stroke speed, and spatial dimensions, through digital tablets or mobile applications. These features are analyzed with machine learning models, including random forests, SVMs, and CNNs, which show significant promise in differentiating dysgraphic handwriting from typical patterns. Kunhoth et al. (2024) [6] proposed a framework using a comprehensive dataset and detailed feature extraction to improve diagnostic accuracy, supporting a more inclusive, device-independent approach. This body of work underscores the importance of real-time data capture and analysis, aiming to make dysgraphia diagnosis accessible, reliable, and non-intrusive.

Recent studies have also examined the use of innovative digital tools for dysgraphia detection, like the SensoGrip smart pen (Bublin et al., 2022) [9], which records real-time handwriting dynamics, including finger pressure and angle. The LSTM-SVM hybrid model in this study achieved over 99% accuracy, outperforming traditional methods that rely solely on static tablet data. Additionally, robotics has been incorporated into dysgraphia detection, as shown by Gouraguine et al. (2023) [13], where a humanoid robot engaged children in handwriting tasks, achieving 91% accuracy with a CNN model. These innovations show the potential for novel technologies to enrich dysgraphia diagnostics, enhancing data quality and classification accuracy.

Deep learning has also shown promise in detecting dysgraphia through automated handwriting analysis. Anand et al. (2023) [23] developed a CNN-based model to classify dysgraphia severity in children, reaching 84% accuracy. Procrustes Analysis for dimensionality reduction, as applied by Lomurno et al. (2023) [14], further validates the value of deep learning in early dysgraphia screening. These methods emphasize both visual and temporal data, as illustrated by Gemelli et al. (2023) [12], who used a ResNet18 model achieving an F1 score nearly on par with human performance.

Optical Character Recognition (OCR) has also advanced dysgraphia research. Hogan (1999) [26] and Rijhwani et al. (2020) [27] utilized OCR to recognize handwriting in minority languages, aiding diagnosis across various linguistic groups. Yin et al. (2019) [28] improved OCR accuracy for Chinese characters using deep learning, achieving a 99.38% rate. Sahu and Sonkusare (2017) [29] reviewed OCR methods for different scripts, highlighting issues like noise and character segmentation that are critical for recognizing dysgraphic handwriting.

Gupta et al. (2023) [17] introduced an AI-based tool to assist dysgraphic individuals with handwriting recognition, spelling correction, and text-to-speech functionality. This approach, integrating CNN-RNN-CTC models with SymSpell for spelling correction, demonstrates the potential of AI in supporting communication for dysgraphic individuals. Ikermane and El Mouatasim (2023) [21] used an artificial neural network to classify dysgraphic handwriting with 96% accuracy, showing the feasibility of tablet-based tools for early screening.

Advancements in dysgraphia detection also include hybrid OCR techniques for cursive scripts, such as Arabic, as studied by Beg et al. (2010) [30]. Hybrid OCR systems combining hardware and software have improved speed and accuracy, especially for scripts with varying character shapes. Such OCR advancements expand access to diagnostic tools across diverse languages.

The literature suggests that gamified mobile apps can aid in dysgraphia diagnosis and intervention. Kariyawasam et al. (2019) [7][10] developed apps using CNNs and SVMs in a gamified environment, engaging children in diagnostic tasks with promising accuracy. These apps offer an accessible, scalable solution for early diagnosis, especially in underserved regions.

Lastly, the need for cross-language adaptability in dysgraphia detection has been underscored by Rashid et al. (2023) [19], who analyzed handwriting quality across languages using classifiers like SVM and Naive Bayes. This work highlights the value of language-independent features in dysgraphia diagnosis, promoting the development of universal screening tools for diverse linguistic settings.

## 3. DATASET

The dataset [Ramlan, S. et al (2024) [31]] used in our research was sourced online and consists of offline handwriting samples from Malaysian schoolchildren, some of whom were identified as potentially having dysgraphia. These samples include sentences written in Malay by primary school students, as well as by children receiving intervention from the Malaysia Dyslexia Association (PDM). Data collection involved having students replicate three sets of provided sentences onto paper forms. These forms were then scanned and converted into digital format to enable further analysis.

Subsequently, image processing techniques were applied to pre-process the samples, converting them to a binary format and switching the colors of the foreground and background, by the data collectors. The handwriting images were classified into two categories: "potential dysgraphia" (PD) and "low potential dysgraphia" (LPD). In total, the dataset contains 249 digital handwriting images from 83 participants, with 114 images in the "potential dysgraphia" (PD) category and 135 in the "low potential dysgraphia" (LPD) category. All samples were prepared in black and white format for consistency in analysis, by the data collectors.

Handwriting affected by dysgraphia, especially among children, displays notable differences from typical handwriting. Common characteristics include irregular spacing between letters and words, uneven letter shapes, inconsistent line alignment, and fluctuating letter sizes [Chung, P. J. et al (2020) [32], Kunhoth, J. et al (2024) [33]]. These characteristics manifest visibly in the handwriting itself. The specific dysgraphia-related handwriting traits can vary across languages, influenced by the native language and its application in the academic context. In this dataset, Malay, the national language of Malaysia, is used [Yamat, H. et al (2014) 34, Mohamad, A. N. A. et al (2022) 35].

The dataset includes three distinct sentence sets that students were instructed to copy onto provided paper forms. The first sentence is simple, containing six frequently used words. The second sentence is longer and more complex than the first. The third sentence is the most intricate, spanning multiple lines and comprising a sequence of connected sentences, as illustrated in Figure 1. The dataset's primary folder contains two subfolders: one for "potential dysgraphia" and the other for "low potential dysgraphia," with all images formatted in binary color.

This dataset consists of 249 images in total, comprising 114 images labeled as "potential dysgraphia" and 135 as "low potential dysgraphia," all saved in JPG format. Labels for images were assigned as 'PD' for potential dysgraphia and 'LPD' for low potential dysgraphia. Handwriting samples with more pronounced dysgraphia symptoms were categorized as potential dysgraphia by experts, while samples showing minimal or no signs of dysgraphia were marked as low potential dysgraphia. Each handwriting sample was scanned from physical paper into digital form, creating a dataset of JPG images, by the data collectors.

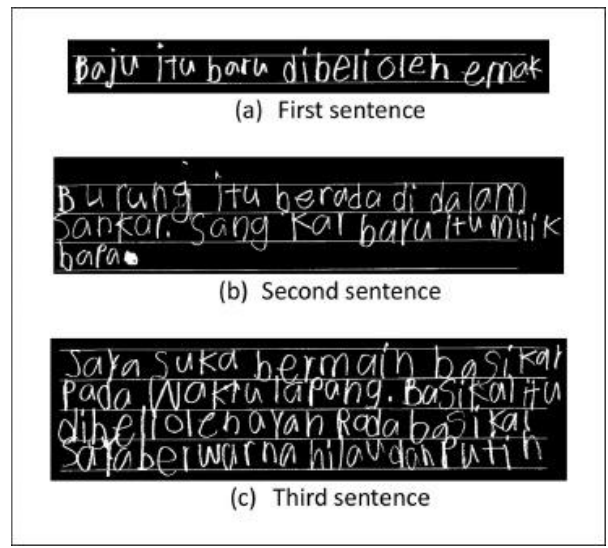

*Figure 1 A sample of the set of sentences used in the data (Ramlan, S. et al (2024) [31])*

## 3.1 Data Preparation for Detection

The dataset images were divided into two main subsets to support the training and validation of the model effectively. An 80-20 split was applied, where 80% of the images were used for training purposes, and the remaining 20% were set aside for validation, ensuring that the model's performance could be evaluated on unseen data. This validation subset allows us to monitor the model's ability to generalize and detect overfitting during the training process.

The preprocessing steps began with resizing each image to a standardized dimension of 224x224 pixels, denoted in the code as img_size = (224, 224). This resizing is crucial for ensuring uniform input size across the dataset, which is particularly important in deep learning models. A consistent input shape allows the model to process images more efficiently and helps prevent issues related to varying dimensions, which could disrupt the model's architecture and performance. Additionally, resizing to 224x224 pixels is common for Convolutional Neural Networks (CNNs), providing a balance between computational efficiency and retention of visual features necessary for dysgraphia detection.

The dataset was further normalized by scaling pixel values to a range between 0 and 1. This normalization was implemented using the rescale parameter in the ImageDataGenerator class from Keras, with the argument set as rescale=1./255. Each pixel's original value, typically ranging from 0 to 255, was divided by 255 to achieve this scaling. Normalization ensures that the model's learning process becomes more stable and converges faster by standardizing input data across images. This step also reduces the impact of lighting variations and other inconsistencies in image quality, which can be particularly important when working with handwriting samples that may differ in scanning conditions and brightness.

To facilitate the data split into training and validation subsets, the validation_split parameter in ImageDataGenerator was set to 0.2, indicating that 20% of the data should be used for validation purposes. This split was controlled within the generator itself, avoiding the need for manual separation of the dataset and ensuring an efficient workflow. Using the subset argument, the data generator could then differentiate between the training and validation subsets, preventing data leakage and maintaining consistent class distribution across both subsets.

In the code, two data generators were created: train_gen and val_gen, representing the training and validation data, respectively. The train_gen generator was configured with the following parameters:

- Directory: The root_dir specifies the directory containing the main dataset folder.
- Target Size: Each image is resized to the specified target dimensions, (224, 224), to match the model's input requirements.
- Batch Size: The batch size for training was set to 32, meaning that the model processes 32 images per iteration before updating weights. This batch size balances computational efficiency and the ability of the model to learn effectively from multiple samples.
- Class Mode: The class_mode parameter was set to 'binary,' as the task involves binary classification between two categories—potential dysgraphia and low potential dysgraphia.
- Subset: The subset parameter for train_gen was specified as 'training,' directing the generator to access only the training images (80% of the dataset).
- Shuffle: Setting shuffle=True ensured that the images in the training subset were presented in a random order, promoting a more robust learning process and preventing the model from learning any sequence biases from the data.

The validation generator, val_gen, used similar parameters but with two slight variations:

- Subset: The subset parameter was set to 'validation,' allowing the generator to access the 20% of images designated for validation.
- Shuffle: Here, shuffle=False was applied, ensuring that the images in the validation set were processed in a consistent order. This configuration is important for repeatable validation results, as shuffling could lead to variability in evaluation metrics across different runs.

By leveraging the ImageDataGenerator with these settings, the dataset was preprocessed effectively, enabling streamlined data loading and real-time image augmentation if required in future model tuning. This approach helped maintain efficient memory usage by loading images in batches, optimizing the training process, and minimizing computational overhead.

## 3.2 Data Preparation for Optical Character Recognition

For the Optical Character Recognition (OCR), preprocessing focused exclusively on images classified as "potential dysgraphia" (PD), as these samples contain handwriting more likely to exhibit the symptoms targeted for recognition and analysis. Since the handwriting images were scanned from notebook paper, they included reference lines that could interfere with accurate character detection and segmentation. To address this, we employed GIMP, an open-source image editing tool, to manually remove these guide lines, ensuring the dataset was optimally prepared for OCR processing.

The line removal process involved the following steps:

1. Importing Images into GIMP: Each scanned image was imported into the GIMP program for editing. This enabled a detailed and customizable approach to removing the lines without impacting the integrity of the handwritten characters.
2. Erasing Guide Lines with the Brush Tool: Using GIMP's brush tool, we manually erased the reference lines present in each image. This meticulous step ensured that the images contained only the handwriting, reducing noise and enhancing the OCR model's ability to focus solely on the written characters.
3. Inverting Colors for Compatibility: Once the guide lines were removed, we adjusted the color scheme by inverting the colors linearly. This step was essential as the character by character data labelling tool was designed to process images with a white background and black text. By inverting the colors, we ensured consistency across the dataset, providing the data labelling tool with the expected input format.

To facilitate character segmentation and labeling, multiple cleaned images were merged into one large composite image. Combining images in this way allowed for more efficient and cohesive segmentation, as it enabled us to label each individual character within a larger context. This consolidated approach helped streamline the dataset, creating a uniform structure for character-level labeling, which is crucial for training deep learning models to recognize and classify individual characters accurately.

Through this comprehensive preprocessing pipeline, we prepared the PD class data for effective character recognition, enhancing the OCR model's potential to learn and generalize from the diverse handwriting samples in the dataset.

# 4. METHODOLOGY

## 4.1 Methodology for Potential Dysgraphia Detection

To detect potential dysgraphia, we experimented with three distinct neural network architectures: a Convolutional Neural Network (CNN), VGG16, and ResNet50. Each model was trained on a dataset of labeled handwriting images for a total of 10 epochs, with the goal of determining which model architecture could most effectively distinguish between the handwriting samples that indicated potential dysgraphia and those that did not.

### 4.1.1 Custom CNN Model Architecture

The custom CNN architecture (depicted in Figure 2) was designed specifically to capture handwriting features linked to dysgraphia symptoms, such as inconsistent letter shapes, varying line spacing, and irregular letter sizes. This model comprised a sequential stack of convolutional layers, followed by max-pooling layers to downsample the feature maps while preserving essential details. The model started with a convolutional layer with 32 filters and a 3x3 kernel, applying ReLU activation. This layer was followed by a max-pooling layer to reduce dimensionality, thereby controlling overfitting while retaining the prominent visual patterns.

Subsequent convolutional layers progressively increased in filter depth (from 64 to 128 filters), each followed by a max-pooling layer to capture increasingly complex patterns within the handwriting data. The model then flattened these feature maps into a dense layer with 128 neurons, activated by ReLU, to compile the information extracted from prior layers into high-level representations. To prevent overfitting, a dropout layer with a dropout rate of 0.5 was added, randomly disabling 50% of neurons during training. Finally, the model ended with a dense layer containing a single neuron with a sigmoid activation function, producing an output value between 0 and 1, corresponding to the binary classification labels of potential dysgraphia and low potential dysgraphia.

### 4.1.2 VGG16 Model Architecture

For the second model, we adapted the VGG16 architecture, a well-established convolutional neural network model known for its high performance in image classification tasks. The pre-trained VGG16 model was used as the base, with weights initialized from the ImageNet dataset to leverage transfer learning. However, to tailor the model to the dysgraphia detection task, the convolutional layers were frozen, meaning the base layers were kept non-trainable to retain the learned features from general image patterns, which are likely beneficial for handwriting data.

A global average pooling layer was appended to the base model to reduce the feature maps, followed by a dense layer with 128 neurons activated by ReLU. The addition of a dropout layer (with a rate of 0.5) provided regularization, ensuring robust training while reducing overfitting. As with the CNN model, a final dense layer with a single neuron using a sigmoid activation function was added for binary classification.

### 4.1.3 ResNet50 Model Architecture

The third model utilized was ResNet50, a residual neural network that introduces skip connections, allowing gradients to flow more effectively through deeper layers. This architecture was chosen due to its strong performance in handling complex image patterns and alleviating the vanishing gradient problem, a challenge in deep networks. As with the VGG16 model, ResNet50 was pre-trained on the ImageNet dataset, with the convolutional layers frozen to maintain the high-level feature representations.

A global average pooling layer was applied to the ResNet50 base, reducing the dimensionality of the feature maps. This was followed by a dense layer with 128 neurons and ReLU activation. To mitigate overfitting, we included a dropout layer with a 50% dropout rate before the final output layer. The last layer was a single-neuron dense layer with a sigmoid activation function, facilitating binary classification.

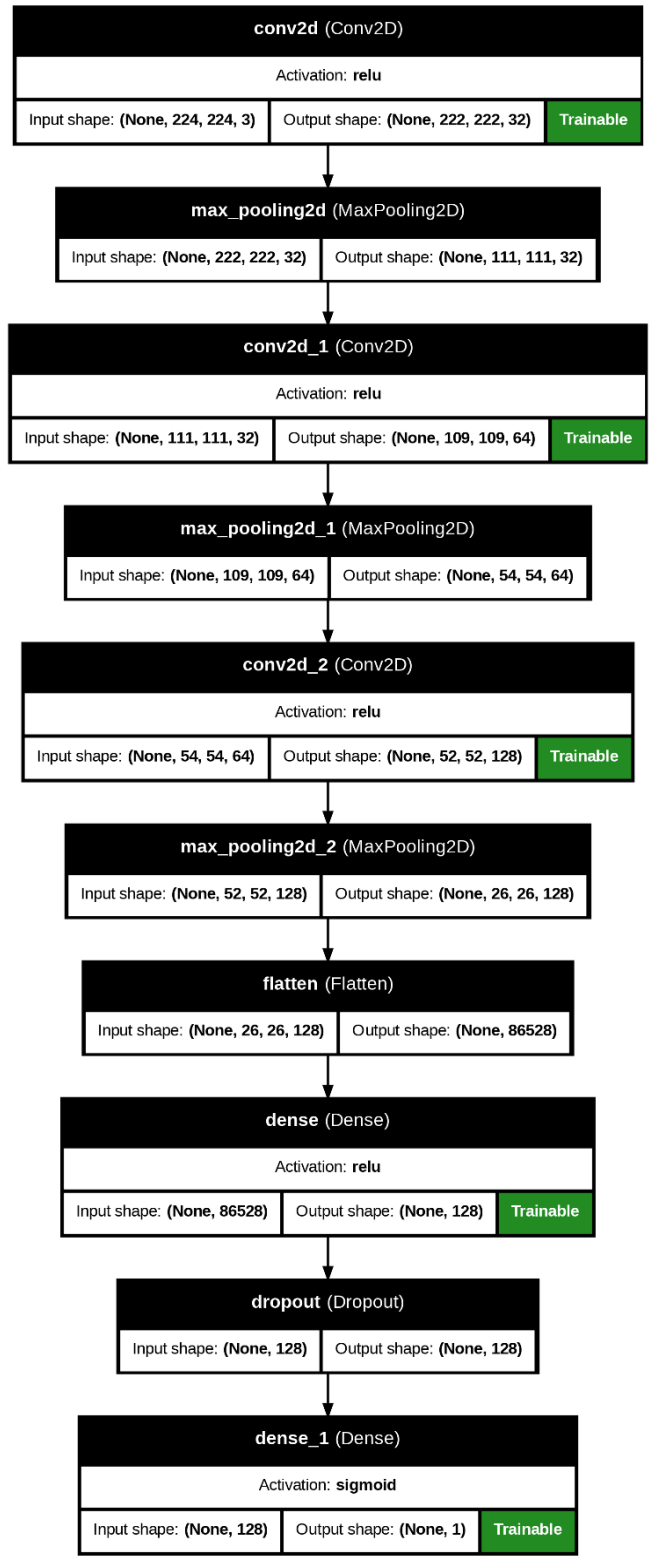

*Figure 2 Custom CNN Model Architecture*

### 4.1.4 Training and Evaluation

Each model was trained using the Adam optimizer, chosen for its adaptive learning rate capabilities and efficiency in training deep learning models. The binary cross-entropy loss function was used to guide the learning process, as it is well-suited for binary classification tasks. Additionally, four performance metrics were monitored: accuracy, precision, recall, and area under the curve (AUC).

To prevent overfitting, early stopping was applied with a patience level of three epochs. This technique monitored validation loss, restoring the model weights from the epoch with the lowest validation loss if no improvement was seen for three consecutive epochs. Each model was evaluated on both the training and validation sets, capturing metrics such as training and validation accuracy, loss, precision, recall, and AUC for a comprehensive understanding of model performance.

### 4.2 Methodology for Optical Character Recognition

To achieve effective Optical Character Recognition (OCR) on handwriting affected by dysgraphia, we developed a customized data labeling tool to facilitate the segmentation and labeling of individual characters from the preprocessed images containing potential dysgraphic symptoms. These images, initially prepared in the data preprocessing step, were combined into larger composite images to streamline the segmentation and labeling process. Using our tool, each character was extracted, resized to a standard size of 40x40 pixels, and adjusted to feature white text on a black background, ensuring compatibility with the OCR model's input expectations.

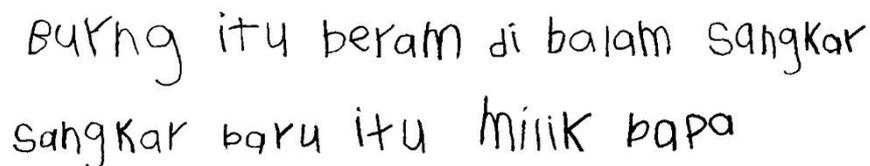

*Figure 3 Sample Image of a sentence after manual preprocessing*

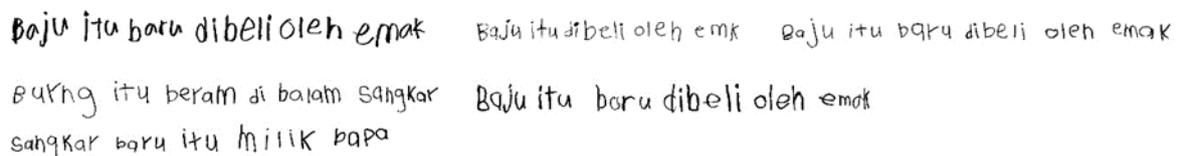

*Figure 4 Image combined with many manually preprocessed images, ready for loading in the data labelling tool*

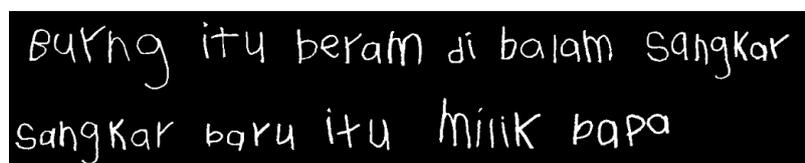

*Figure 5 Part of Sample image ready for character segmentation (inside data labelling tool)*

### 4.2.1 Character Segmentation through Contouring Technique

The core segmentation process relied on contouring, a technique that identifies the boundaries of distinct regions within an image based on pixel intensity. Contours help identify shapes within the image by detecting and outlining continuous curves or boundaries, which is particularly useful for segmenting handwritten characters. To achieve this, thresholds were applied to separate background and foreground elements effectively. In our case, the color inversion applied during preprocessing enabled the tool to recognize and isolate white handwriting on a black background as individual contours.

### 4.2.2 Addressing Issues with Traditional Contouring Techniques

Challenges in segmenting dysgraphic handwriting necessitated multiple iterations and adjustments to the general contouring process. Dysgraphic handwriting often includes irregularly shaped and spaced letters, which presented unique obstacles in character segmentation. Several problems and solutions emerged through iterative refinements, as outlined below.

1. Separation of Joint Letters: Traditional contouring techniques frequently failed to segment characters that were written close together or were improperly spaced, which is common in dysgraphic handwriting. This was particularly problematic when two or more characters overlapped or were connected by improper spacing. In these instances, the tool identified the entire connected sequence as one single bounding box, failing to differentiate between individual letters.

   To overcome this, we introduced a thresholding mechanism based on the median width of all segmented characters. By calculating the median width, we established a baseline for typical character width and set a threshold at 1.5 times this median value. When a bounding box's width exceeded this threshold, the tool estimated the number of characters it likely contained. The bounding box was then divided horizontally according to this estimation, effectively creating individual bounding boxes for each character within the sequence. Any extraneous whitespace within these new boxes was then removed, providing cleaner segmentation.

2. Identification of Dotted Characters ('i' and 'j'): A unique challenge involved correctly identifying letters like 'i' and 'j,' where the dot is physically separated from the main stem of the letter. Contouring often misclassified the dot as an independent character, which caused segmentation errors. In dysgraphic handwriting, the dot is often misaligned, positioned to the left or right of the stem due to irregular letter spacing.

To address this, a height threshold was implemented to identify potential dots above shorter characters. For instances where a potential dot was identified, a width threshold was used to search downwards along the image width for a corresponding bounding box, likely containing the main stem of the letter. Given the characteristic dimensions of 'i' and 'j,' we could identify the correct bounding box containing the stem by selecting the box with the minimum width. Once identified, the two boxes (dot and stem) were merged to form a single bounding box representing the complete character. Any remaining whitespace within these merged boxes was removed to further improve segmentation accuracy.

The segmented characters, now accurately bounded and refined, were resized to 40x40 pixels and converted to a white-on-black color scheme. This final format provided a uniform dataset for training the OCR model, ensuring that each character's features were standardized, facilitating improved model learning and generalization.

Through these iterative refinements and optimizations, our OCR preprocessing pipeline adapted effectively to the unique challenges posed by dysgraphic handwriting, yielding well-defined and labeled characters suitable for deep learning model training.

After preprocessing, the segmented letters dataset was prepared for training. Few Characters were randomly rotated within the range of -15 degrees to +15 degrees and randomly blurred for data augmentation to make the generalisation property better. A custom CNN model was developed and trained individually on each character in the dataset, with a 20% test split to evaluate model performance. This CNN model was trained for four epochs, allowing it to learn distinctive character features efficiently and generalize effectively. Once trained, the model was used to predict character sequences within each image, implementing a combination of techniques to achieve accurate OCR results.

### 4.2.3 Techniques Used for OCR Output Extraction

1. Segmentation and Bounding Box Detection:

   The process began by extracting characters and their respective bounding boxes from each input image. Using contour-based segmentation, each character was isolated by detecting boundaries based on pixel intensity thresholds. These contours defined bounding boxes that encompassed individual characters, enabling efficient localization. After extracting each character image, its coordinates were recorded to facilitate subsequent ordering and grouping.

2. Row Detection and Character Grouping:

    With the character bounding boxes identified, the next step was to group characters into rows, mimicking the natural arrangement of lines in text. Bounding boxes were first sorted by their y-coordinates to differentiate rows. To define a separation threshold between rows, the median height of all bounding boxes was calculated, then multiplied by a factor (set at 1.5) to adjust for potential dysgraphic variances. Bounding boxes exceeding this threshold in y-coordinate were grouped into separate rows, effectively aligning text segments into distinct lines.

3. Left-to-Right Character Sorting within Rows:

    Once characters were grouped by rows, each row was sorted based on x-coordinates to ensure left-to-right order. This step preserved the natural flow of the text and ensured accurate sequencing of characters within each row.

4. Space Detection and Handling:

    Dysgraphic handwriting often presents irregular spacing between characters, making it challenging to distinguish intentional spaces between words. To address this, a spacing threshold was applied based on the median width of bounding boxes. If the horizontal gap between consecutive bounding boxes exceeded this threshold, a space was inserted in the text output, effectively recognizing word boundaries.

5. Identification and Correction of Dotted Characters ('i' and 'j'):

    The segmentation process faced challenges in identifying dotted characters like 'i' and 'j,' as their dots are often isolated from the main character stems in dysgraphic handwriting. To address this, height and width thresholds were introduced. The algorithm searched for small bounding boxes above short stems, then identified corresponding bounding boxes within a horizontal search range, assuming these contained the stems. By comparing bounding box dimensions, the smallest box was identified as the primary character component. This bounding box and the dot box were merged to represent a complete character, followed by the removal of unnecessary whitespace within these merged boxes.

6. Character Recognition and Text Annotation:

   For each character image, preprocessing included resizing to 40x40 pixels and normalization before feeding it into the trained CNN model. The model predicted each character based on probabilities for each class, assigning the label with the highest probability as the recognized character. Labels were then mapped to corresponding characters (a–z) and arranged into sentences.

7. Bounding Box and Character Label Annotation:

   For visual verification, the original grayscale images were converted to RGB format to allow overlaying of character labels and bounding boxes in the output image. Each character's bounding box was outlined, and the recognized character was annotated above it. Additionally, each character was assigned an index (starting from 1) for sequential tracking, which helped validate the reading order.

8. Sentence Construction and Final Output:

   The characters within each row were compiled to construct complete sentences. These sentences were appended line-by-line to form the final OCR output, accurately reflecting the segmented text. The resulting annotated image was then saved, showing each character prediction and bounding box, allowing for a complete and visually comprehensible OCR output.

This methodology enabled the OCR system to handle the unique characteristics of potentially dysgraphic handwriting effectively, resulting in robust recognition outputs that account for both irregular letter spacing and character-specific challenges.

## 5. RESULTS AND DISCUSSION

This section presents the outcomes and analysis of our dysgraphia detection and OCR pipelines, highlighting both the effectiveness of the models used and the challenges encountered. The results are evaluated based on several key performance metrics, which provide insights into the model's ability to generalize and accurately classify dysgraphic handwriting.

## 5.1 Evaluation Metrics

For the dysgraphia detection models, the primary evaluation metrics included:

- Accuracy: The ratio of correctly classified instances to the total instances, indicating the overall correctness of the model.
- Loss: Represents the model's error in prediction, with lower values indicating a closer alignment between the model's predictions and the actual data. In binary classification, binary cross-entropy was used as the loss function.
- Precision: The proportion of true positives to the sum of true positives and false positives. Precision indicates how accurately the model identifies dysgraphic handwriting when it makes a positive prediction.
- Recall: The proportion of true positives to the sum of true positives and false negatives, reflecting the model's ability to capture all instances of dysgraphic handwriting.
- Area Under the Curve (AUC): Measures the model's ability to distinguish between classes. A higher AUC value indicates better discrimination between potential dysgraphic and non-dysgraphic samples.

For the OCR pipeline, character-level accuracy was used to evaluate how accurately the model predicted each character in the text output. This was calculated by dividing the number of correctly identified characters by the total number of characters in the actual sentence.

The results for both dysgraphia detection and OCR are discussed in detail below, including model-specific observations and implications.

## 5.2 Potential Dysgraphia Detection Results

The results of the potential dysgraphia detection models are summarized in Table 1 for the training phase and Table 2 for the testing phase. Three models were tested: CNN, VGG16, and ResNet50, each evaluated on the metrics described above. Figures 6, 7, and 8 illustrate the accuracy and loss progression over the training epochs for CNN, VGG16, and ResNet50, respectively.

*Table 1 Training Phase Results of the Potential Dysgraphia Detection Models*

| Model | Train Accuracy | Train Loss | Train Precision | Train Recall | Train AUC |
| --- | --- | --- | --- | --- | --- |
| CNN | 0.99 | 0.0604 | 0.978723 | 1 | 1 |
| VGG16 | 0.83 | 0.445758 | 0.871795 | 0.73913 | 0.901973 |
| ResNet50 | 0.83 | 0.445529 | 0.845238 | 0.771739 | 0.87963 |

*Table 2 Testing Phase Results of the Potential Dysgraphia Detection Models*

| Model | Test Accuracy | Test Loss | Test Precision | Test Recall | Test AUC |
|---|---|---|---|---|---|
| CNN | 0.918367 | 0.261982 | 0.95 | 0.863636 | 0.955387 |
| VGG16 | 0.836735 | 0.460559 | 1 | 0.636364 | 0.944444 |
| ResNet50 | 0.836735 | 0.445183 | 1 | 0.636364 | 0.927609 |

During training, the custom CNN model achieved the highest performance, with a training accuracy of 99% and validation accuracy of 91.8%, as well as superior precision, recall, and AUC scores. Its validation loss was notably lower than the other models, at 0.26. The CNN model's high accuracy and low loss values suggest that it effectively learned the distinguishing features of dysgraphic handwriting, allowing it to generalize well on validation data. This high performance was also evident in the test phase, where the CNN achieved a test accuracy of 91.8%, with high precision and recall, underscoring its robustness in distinguishing handwriting associated with dysgraphia.

In contrast, the VGG16 and ResNet50 models both achieved similar training and validation accuracies of around 83%, with validation losses higher than the CNN model. These pre-trained architectures, though powerful, may not have been as optimized for the specific handwriting characteristics unique to dysgraphia, possibly due to limitations in transfer learning when applied to a niche dataset. In testing, both VGG16 and ResNet50 maintained validation accuracies of 83.6% and achieved perfect precision (1.0), but had comparatively lower recall scores (63.6%), indicating they were more conservative in their predictions, often missing cases of dysgraphia.

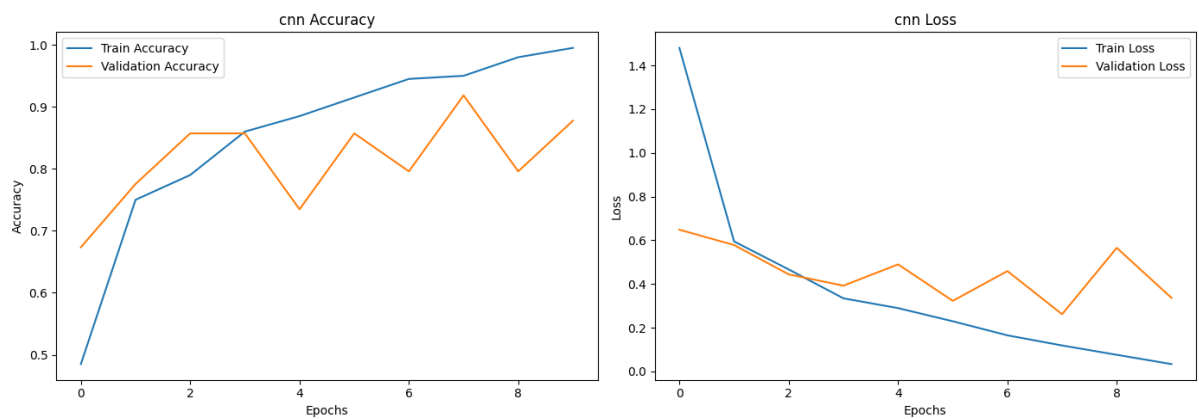

*Figure 6 Accuracy and Loss of CNN model for potential dysgraphia detection*

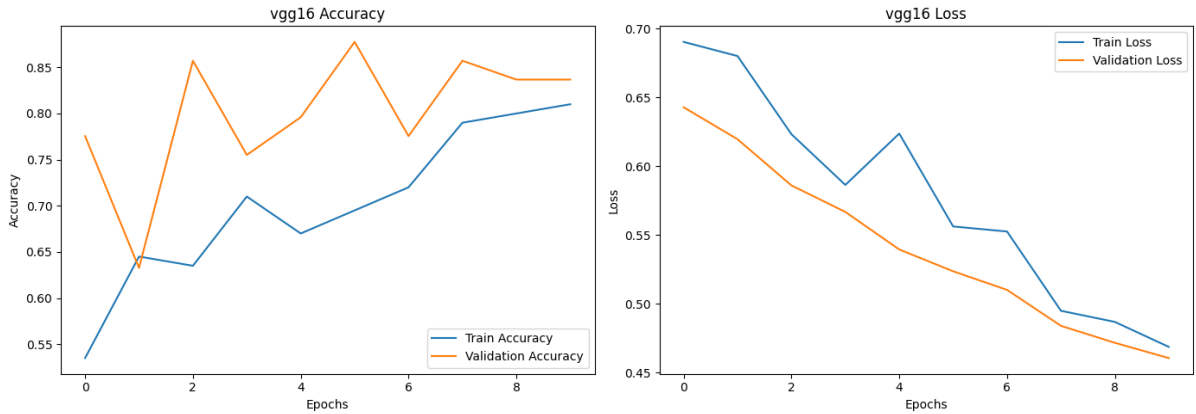

*Figure 7 Accuracy and Loss of VGG16 model for potential dysgraphia detection*

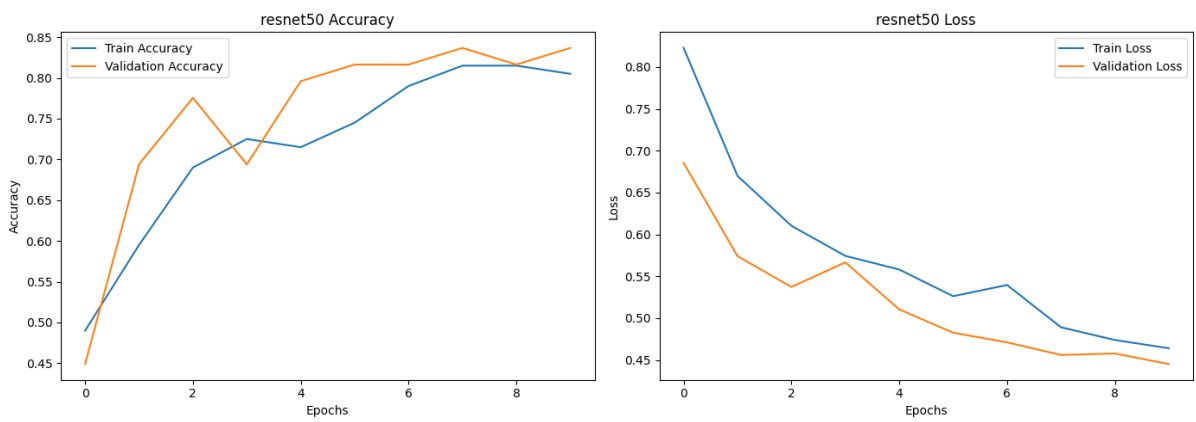

*Figure 8  Accuracy and Loss of ResNet50 model for potential dysgraphia detection*

These results suggest that while all three models perform well in detecting dysgraphia, the custom CNN model demonstrates superior performance, likely due to its tailored architecture optimized for this specific task. However, the performance of VGG16 and ResNet50 also illustrates the potential of transfer learning, albeit with certain limitations when applied to smaller, specialized datasets.

## 5.3 OCR Pipeline Results

The OCR component, focused exclusively on segmenting and recognizing characters from handwriting classified as potential dysgraphia, yielded mixed results in translating dysgraphic handwriting into text.

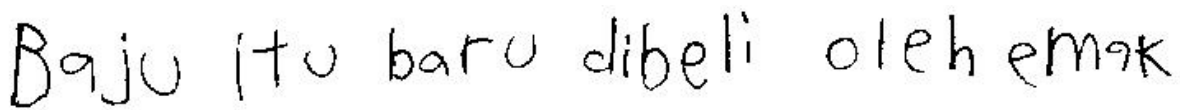

*Figure 9 Sample Testing image for OCR Pipeline*

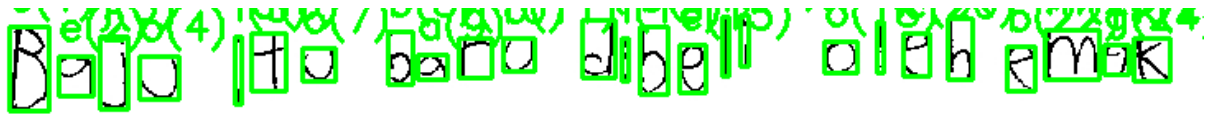

*Figure 10 Annotated output of the sample Testing image for OCR Pipeline*

To evaluate OCR accuracy, we calculated the character-level accuracy as the number of correctly identified characters divided by the total number of characters in the sentence. For one of the test images, shown in Figure 9, the model output was "eejo ido baro dibeli oleb bmgk," (shown in Figure 10) while the actual sentence was "baju itu baru dibeli oleh emak." In this case, the OCR model achieved an accuracy of 10 out of 23 characters correctly identified, resulting in an accuracy of approximately 43.5%. On all testing images, the OCR accuracy ranged within ±5% of this value, indicating variability in performance depending on the quality and spacing of individual handwriting samples.

The OCR output demonstrates both the potential and limitations of the current model. While the OCR pipeline was able to capture certain characters accurately, many characters were incorrectly identified or transposed, likely due to the irregular spacing, inconsistent character formation, and complex segmentation requirements characteristic of dysgraphic handwriting. These results suggest that while the model can provide some level of character recognition, additional improvements are needed, particularly in handling inconsistent letter spacing and separation between joint letters.

## 5.4 Discussion

The dysgraphia detection and OCR results underscore the potential of deep learning in identifying and interpreting dysgraphic handwriting, highlighting specific models' suitability for clinical and educational applications. In an educational context, the dysgraphia detection model can serve as a diagnostic aid for teachers and clinicians, providing early indications of dysgraphia and enabling timely intervention. Accurate OCR capabilities could further facilitate tracking a child's handwriting progress over time, offering valuable insights into improvements or challenges in fine motor skills.

However, challenges persist, particularly concerning dataset limitations. A more extensive and diverse dataset encompassing varied handwriting styles, including more samples from different age groups, could further strengthen model robustness. Additionally, the nuances of dysgraphic handwriting, such as inconsistent letter spacing and positioning, suggest a need for tailored model architectures or specialized preprocessing techniques in OCR applications.

Overall, this research provides a foundation for developing automated tools to detect and analyze dysgraphic handwriting. Future work may include expanding the dataset, refining model architectures, and exploring hybrid approaches to enhance detection and OCR accuracy in dysgraphia-related applications.

# 6. CONCLUSION AND FUTURE WORK

This study explored the application of deep learning models for detecting dysgraphia and performing optical character recognition (OCR) on handwriting samples from children with potential dysgraphic symptoms. Key achievements include the successful development and evaluation of a custom Convolutional Neural Network (CNN) that outperformed pre-trained architectures like VGG16 and ResNet50 in dysgraphia detection. The CNN model demonstrated high accuracy and robustness in distinguishing handwriting associated with dysgraphia, achieving a test accuracy of 91.8%, with high precision and recall metrics. Additionally, the study proposed a customized OCR pipeline tailored for dysgraphic handwriting, which showed promise in segmenting and recognizing characters, despite challenges arising from irregular spacing and inconsistent letter formation characteristic of dysgraphia.

The OCR pipeline provided a character-level accuracy of approximately 43.5%, reflecting the complexity of working with dysgraphic handwriting. While the OCR model achieved partial success, the character recognition process was hindered by the limitations of the dataset and the nuanced nature of dysgraphic writing. Nevertheless, these results underscore the potential of using deep learning to assist educators and clinicians in assessing dysgraphia and monitoring handwriting improvements over time.

Future work could enhance the models through:

1. Dataset Expansion: Collecting a larger, more diverse dataset across different age groups and dysgraphia severity levels to improve model generalization and robustness.
2. OCR Pipeline Refinement: Developing advanced segmentation techniques and incorporating post-processing corrections could improve character recognition accuracy, especially for challenging handwriting.
3. Age-Specific Model Adaptation: Tailoring models for age-specific handwriting patterns could improve accuracy in detecting dysgraphia across various age groups.
4. Real-World Application: Creating a user-friendly tool based on these models could support dysgraphia assessment in educational and clinical settings, enabling progress tracking and personalized feedback.

In summary, this research lays a foundation for leveraging deep learning in dysgraphia detection and handwriting analysis. With continued refinements, larger datasets, and application-specific adaptations, this approach holds promise for making significant contributions to the fields of educational technology and assistive diagnostics in dysgraphia.